\documentclass[11pt, letterpaper]{article}

\usepackage[
  top=0.85in, bottom=0.9in,
  left=0.85in, right=0.85in
]{geometry}

\usepackage[utf8]{inputenc}
\usepackage[T1]{fontenc}
\usepackage{microtype}

\usepackage{amsmath, amssymb}
\usepackage{nicefrac}

\usepackage[table,dvipsnames]{xcolor}
\definecolor{titleBlue}{HTML}{1A4F8A}       
\definecolor{linkBlue}{HTML}{2563EB}        
\definecolor{accentBlue}{HTML}{1D4ED8}      
\definecolor{metaGray}{HTML}{555555}        
\definecolor{ruleGray}{HTML}{CCCCCC}        
\definecolor{kwBg}{HTML}{EFF6FF}            
\definecolor{kwBorder}{HTML}{BFDBFE}        
\definecolor{abstractBg}{HTML}{F8FAFC}      
\definecolor{tableHeaderBg}{HTML}{F3FBF5}   
\definecolor{cheapRowBg}{HTML}{F3F8FF}      
\definecolor{expensiveRowBg}{HTML}{FFEBEE}  
\definecolor{reasoningRowBg}{HTML}{FFF4F0}  
\definecolor{cheapLabelBg}{HTML}{DCEEFF}    
\definecolor{expensiveLabelBg}{HTML}{FFCCCB}
\definecolor{reasoningLabelBg}{HTML}{FFE2D8}

\usepackage[
  colorlinks=true,
  linkcolor=linkBlue,
  citecolor=linkBlue,
  urlcolor=linkBlue,
  pdftitle={Absurd World: A Simple Yet Powerful Method to Absurdify the Real-world for Probing LLM Reasoning Capabilities},
  pdfauthor={Authors},
]{hyperref}

\usepackage{graphicx}
\usepackage{booktabs}
\usepackage{multirow}
\usepackage{float}
\usepackage{placeins}
\usepackage{subcaption}
\usepackage{fontawesome5}
\usepackage{tabularx}

\usepackage{enumitem}
\usepackage{parskip}
\usepackage{array}
\usepackage{tabularx}
\usepackage{calc}

\usepackage[most]{tcolorbox}
\tcbuselibrary{breakable, skins}

\usepackage{fancyhdr}
\usepackage{adjustbox}
\newcommand{\githublogo}{%
  \adjustbox{valign=c}{\includegraphics[height=3.25ex]{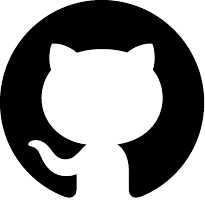}}}

\newcommand{\hflogo}{%
  \adjustbox{valign=c}{\includegraphics[height=5.55ex]{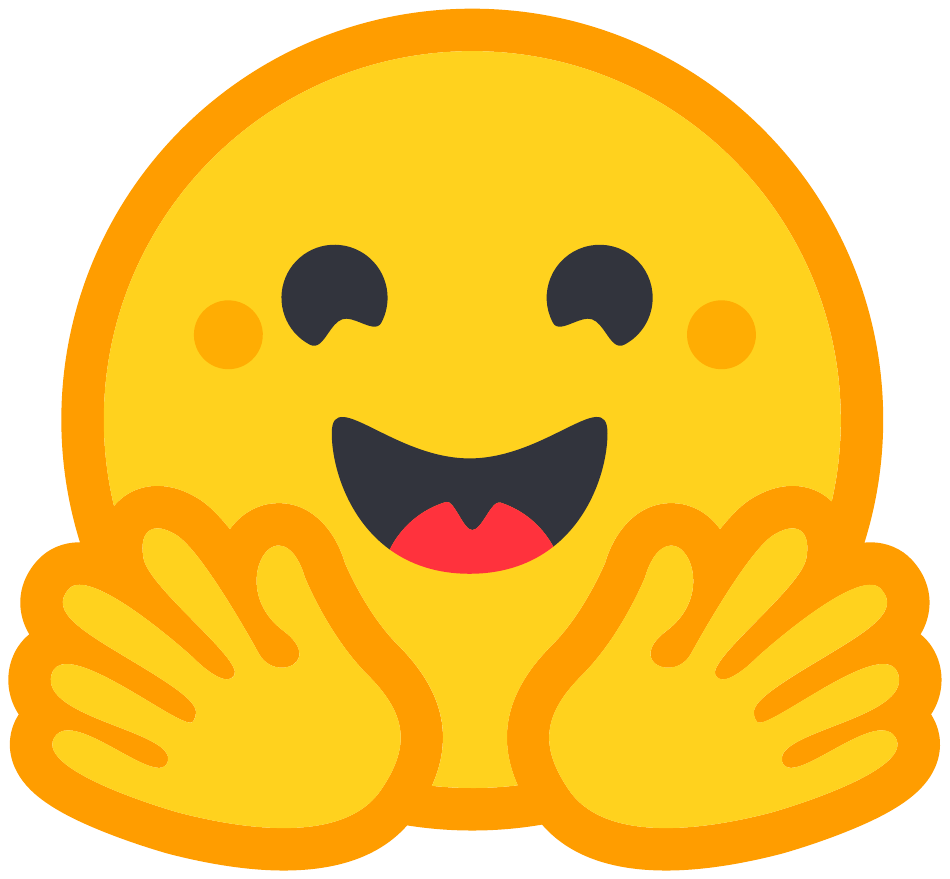}}}

\usepackage{titlesec}
\titleformat{\section}
  {\normalsize\bfseries\color{accentBlue}}
  {\thesection}{0.5em}{}
\titleformat{\subsection}
  {\normalsize\bfseries\color{accentBlue}}
  {\thesubsection}{0.5em}{}
\titleformat{\subsubsection}
  {\small\bfseries\itshape}
  {\thesubsubsection}{0.5em}{}
\titlespacing*{\section}{0pt}{12pt}{5pt}
\titlespacing*{\subsection}{0pt}{9pt}{3pt}

\usepackage[numbers,compress]{natbib}
\newtcolorbox{promptbox}[1]{
  enhanced, breakable,
  colback=abstractBg, colframe=accentBlue!65!black,
  arc=4pt, boxrule=0.8pt,
  left=8pt, right=8pt, top=6pt, bottom=6pt,
  fonttitle=\bfseries\small\color{white}, coltitle=white,
  title=#1,
  attach boxed title to top left={xshift=8pt, yshift=-\tcboxedtitleheight/2},
  boxed title style={colback=titleBlue, arc=2pt, boxrule=0pt, left=5pt, right=5pt}
}

\newcommand{\inlinebadge}[3]{%
  \tcbox[on line, arc=3pt,
    colback=#1, colframe=#1,
    boxrule=0pt, boxsep=0pt,
    left=4pt, right=4pt, top=2pt, bottom=2pt,
    fontupper=\scriptsize\bfseries\color{#2}]{#3}}

\newcommand{\ghbadge}[1]{\href{#1}{\inlinebadge{black}{white}{GitHub}}}
\newcommand{\hfbadge}[1]{\href{#1}{\inlinebadge{orange!80!yellow}{white}{Hugging Face}}}
\newcommand{\axvbadge}[1]{\href{#1}{\inlinebadge{red!70!black}{white}{arXiv}}}

\newcommand{\kw}[1]{%
  \tcbox[on line, arc=3pt,
    colback=kwBg, colframe=kwBorder,
    boxrule=0.5pt, boxsep=0pt,
    left=4pt, right=4pt, top=2pt, bottom=2pt,
    fontupper=\scriptsize\color{accentBlue}]{#1}\,%
}

\begin{document}

\begin{center}
  {\LARGE\bfseries\color{titleBlue}
    Absurd World: A Simple Yet Powerful Method \\[3pt]
    to Absurdify the Real-world for Probing \\[3pt]
    LLM Reasoning Capabilities
  \par}
\end{center}
  \vspace{10pt}

\begin{center}
  {\small
    \textbf{Ryan Albright}\textsuperscript{1,*},\enspace
    \textbf{Golam Md Muktadir}\textsuperscript{},\enspace
    \textbf{Zarif Ikram}\textsuperscript{3,*},\enspace
    \textbf{S M Jubaer}\textsuperscript{4},\enspace
    \textbf{Mehrab Hossain}\textsuperscript{5}, \enspace
    \textbf{Dianbo Liu}\textsuperscript{6}
  }
\end{center}

\vspace{5pt}

\begin{center}
  {\footnotesize\color{metaGray}
    \textsuperscript{1}The Nueva School, San Mateo, USA\enspace·\enspace
    \textsuperscript{3}University of Southern California\enspace·\enspace
    \textsuperscript{4}Notre Dame College, Dhaka, Bangladesh\enspace·\enspace
    \textsuperscript{5}Arizona State University\enspace·\enspace
    \textsuperscript{6}National University of Singapore \\[2pt]
    \textsuperscript{*}Corresponding Authors \\[2pt]
    \texttt{ryaalbr@nuevaschool.org}\enspace·\enspace
    \texttt{zikram@usc.edu}
  }
  
\end{center}

\vspace{7pt}

{\color{titleBlue}\rule{\linewidth}{2pt}}

\vspace{6pt}

\begin{tcolorbox}[
  enhanced, breakable,
  colback=abstractBg,
  colframe=accentBlue!40,
  arc=6pt, boxrule=0.7pt,
  left=12pt, right=12pt, top=9pt, bottom=9pt,
  shadow={1.5pt}{-1.5pt}{0pt}{black!8}
]
  \begin{center}
    {\small\bfseries\color{accentBlue}\MakeUppercase{Abstract}}
  \end{center}
  \vspace{3pt}
  {\small
    While extremely powerful and versatile at various tasks, the thinking capabilities of large
    language models (LLMs) are often put under scrutiny as they sometimes fail to solve problems
    that humans can systematically solve. However, recent literature focuses on breaking LLM
    reasoning with increasingly complex problems, and whether an LLM is robust in simple logical
    reasoning remains underexplored. This paper proposes \textbf{Absurd World}, a benchmarking
    framework, to test LLMs against altered realism, where scenarios are logically coherent,
    and humans can easily solve the tasks. Absurd World breaks a real-world model into
    \emph{symbols}, \emph{actions}, \emph{sequences}, and \emph{events}, which are automatically
    altered to create absurd worlds where the logic to solve the tasks remains the same.
    It evaluates a large collection of models with simple and advanced prompting techniques,
    and proves that it is an effective tool to determine LLMs' ability to think logically,
    ignoring the patterns learned from the real world. One can use this framework to extensively
    test an LLM against a real-world problem to verify whether the LLM's reasoning capability
    is robust against variations of the task.
  }
\end{tcolorbox}

\vspace{4pt}

\noindent{\small\textbf{Keywords:}}\enspace
\kw{LLM Reasoning}
\kw{Benchmarking}
\kw{Absurd World}
\kw{Rule Following}
\kw{In-context Learning}
\kw{Agent Evaluation}

\vspace{4pt}
{\color{ruleGray}\rule{\linewidth}{0.6pt}}


\noindent
\begin{tabular}{@{}ll@{}}
{\small\textbf{\color{metaGray}\githublogo\ GitHub:}} &
{\small\href{https://github.com/ju-baer/Absurd-World}
{\texttt{https://github.com/ju-baer/Absurd-World}}}
\\[2pt]
{\small\textbf{\color{metaGray}\hflogo\ Hugging Face:}} &
{\small\href{https://huggingface.co/datasets/jub-aer/Absurd-World}
{\texttt{https://huggingface.co/datasets/jub-aer/Absurd-World}}}
\end{tabular}

\vspace{4pt}
{\color{ruleGray}\rule{\linewidth}{0.6pt}}
\vspace{8pt}

\section{Introduction}



Recent trends in the AI industry point towards the deployment and use of AI agents. However, expectations for Agentic AI have been hampered by their real-world performance. Researchers have found that LLM agents only achieve a single-turn success rate of 58\% in CRMArena-Pro, a benchmark for the evaluation of LLM agents in business settings~\citep{huang2025crmarenaproholisticassessmentllm}. Such failure is a result of flaws in these models' reasoning ability.


Various methods and datasets aim at testing LLM reasoning abilities. Often, the methods include tasks that are either complex or require a long memory horizon. The tasks are challenging for humans, even when equipped with additional tools such as a pen and paper or a calculator. The ``Tower of Hanoi" game is such an example from \cite{shojaee2025illusion}. But it is important to ensure that LLMs that are becoming our allies in daily life perform reliably in simple tasks in a simple world.

This work examines whether LLMs can reliably solve problems that humans can solve using only limited memory, which we refer to as ``small tasks". Such tasks require models to retain and correctly apply a small amount of task-specific information rather than rely on extensive context or long reasoning traces. If a model fails in this regime, it is difficult to trust it on larger tasks that impose even greater memory demands and deal with diverse concepts. Importantly, the goal is not to explain why LLMs fail at an underlying level, but to systematically identify when they fail and which models are better suited for specific task settings using prompts only. By focusing on small, fully specified tasks, this work introduces a benchmark that isolates basic rule execution and enables automated, controlled comparison across models, providing a practical tool for assessing reliability and guiding model selection and deployment.

Dataset-oriented approaches are promising but are very costly to ensure diversity, variation, and quality, even when the tasks are small. The proposed approach is also a very effective tool to augment existing datasets automatically to test the underlying reasoning capabilities to solve similar tasks.

The rest of the paper is organized as follows: Section \ref{sec:the-absurd-world} describes the Absurd World - the approach to generating test scenarios from a real-world model and task. It shows an example of a penalty-style soccer shootout, and describes how to create multiple ``absurd" world models out of it. Section \ref{sec:related} reviews related work on testing LLM reasoning abilities, aiming at categorizing based on the approaches, rather than specific datasets. It also differentiates Absurd World from existing approaches. Section \ref{sec:experimental-settings} describes the methodology for testing LLMs to answer questions relating to the example absurd world. Then, Section \ref{sec:results} reports the results. Most notable findings includes (1) non-reasoning models are more sensitive than reasoning models to rule changes, (2) few-shot prompting degraded performance from zero-shot prompting, and (3) expensive non-reasoning models performed significantly worse than cheap non-reasoning models. Our suggestions for future work are described in Section \ref{sec:future-work}.


\section{The Absurd World}\label{sec:the-absurd-world}

This work proposes Absurd World, a controllable benchmark designed to isolate a previously unmeasured failure mode in large language models: whether a model is executing the stated rules, or leaning on pretrained world priors. Unlike most of the existing reasoning benchmarks that increase difficulty by adding more steps, longer contexts, or more knowledge, the tasks keep the underlying algorithmic structure fixed and computationally trivial—solvable by simple counting or bookkeeping while systematically perturbing the semantics of the world through rule inversions. This design cleanly decouples reasoning complexity from world familiarity, allowing us to directly test whether models follow explicit symbolic instructions or instead default to prior-consistent world models even when those priors are contradicted by the prompt.

To test basic thinking capabilities, this work proposes a novel approach: logical inference in a small, absurd world model. In an absurd world, there are actors, objects, interactions, and goals that are as coherent as the real world; however, a few, if not all, elements from an absurd world do not exist or make sense in the real world. Defining such worlds and stories in them is a difficult and costly endeavor. To make it feasible, this work also proposes a simple yet effective method to re-purpose real-world models into absurd world models. 

\begin{figure}[H]
    \centering
    \includegraphics[width=1\linewidth]{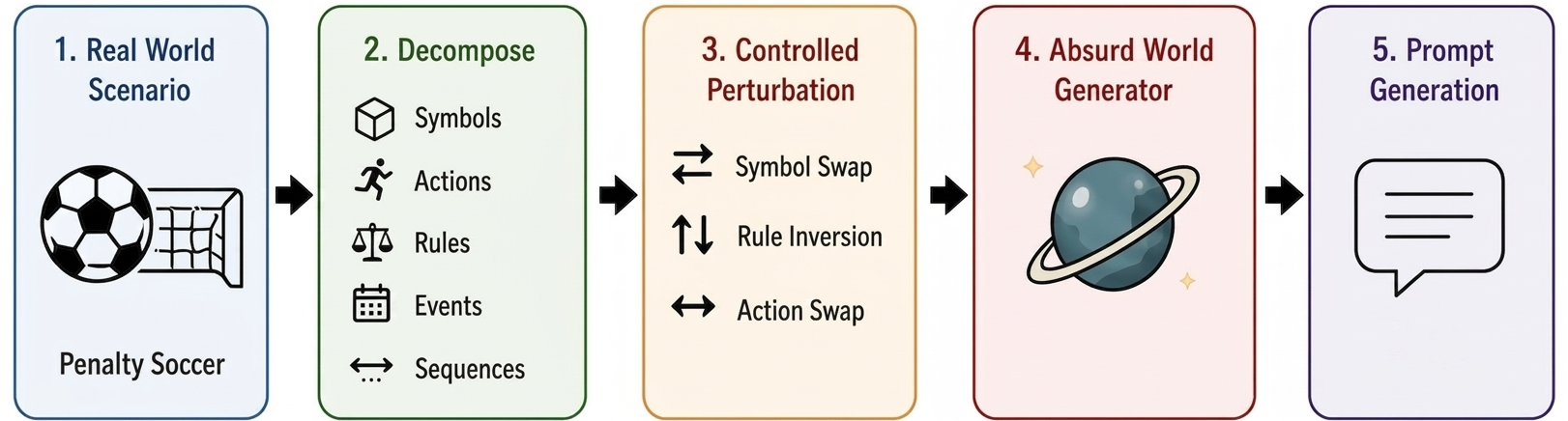} 
    \caption{Overview of Absurd World. Real-world tasks are decomposed into symbolic components, systematically perturbed, transformed into absurd worlds, and used to evaluate LLM rule execution.}
    \label{fig:absurd-world}
\end{figure}

Absurd World de-structures a world model into symbols, actions, and rules, and the scenarios into events and sequences. Symbols represent actors and objects that have a specific role in the world. Actions are a special category of symbols which define the interactions and produce events in the world. Rules define the constraints of the world and determine the consequences of actions. Events contain interaction between symbols using some action and may have numeric elements. Sequence are the ordering of the events in time. A slight change in the ordering can change the entire scenario. Once de-structured, a real-world scenario can be turned into an absurd one by (a) swapping the roles of the symbols, or modifying (b) action symbols, (c) rules, or (d) reordering the sequence of events. Figure~\ref{fig:absurd-world} provides an overview of the full pipeline, from decomposing a real-world scenario into symbolic primitives to generating absurd variants and evaluating model behavior under controlled perturbations.

To validate this approach, we defined a game of soccer, called ``Absurd Soccer" with numerous absurd rulesets that differ from how soccer is usually played. These rulesets are then tested against a set of models, tasking them to determine the outcomes of soccer matches.

\subsection{The Game}\label{the-game}


The game is a simple and familiar real-world one: a penalty-style soccer shootout. Two teams, Team A and Team B, take turns shooting a ball at a net. Each team shoots five times. Every shot is either a \emph{hit} (the ball goes into the net) or a \emph{miss} (the ball does not). Each hit gives 1 point, each miss gives 0 points, and the team with the higher total score at the end wins. Humans can easily read a short text commentary for such a shootout, count the hits for each team, and determine the winner.

To turn this real game into a modifiable world model, it is broken down into three basic components:
\begin{itemize}
    \item \textbf{Symbols}: the main objects and entities in the game, such as the players, the ball, and the net.
    \item \textbf{Actions}: what those symbols can do, such as a team shooting the ball and either hitting or missing the net.
    \item \textbf{Rules}: how the game state changes and how a winner is decided, for example, ``each hit adds 1 point'' and ``the team with the most points wins.''
\end{itemize}

Once the game is written in terms of symbols, actions, and rules, one can create new versions by changing one or more of these parts. This produces a family of absurd but internally consistent games that remain easy for humans to understand but no longer align with standard real-world expectations. For example, one can:
\begin{itemize}
    \item Change the rules so that the team with the fewest points wins instead of the most.
    \item Change the actions so that missing the net earns points instead of hitting the net.
    \item Change symbols, such as swapping the positions of the ball and the net in the description while keeping the logical consequences defined by the rules.
\end{itemize}

\begin{figure}
    \centering
    \includegraphics[width=0.75\linewidth]{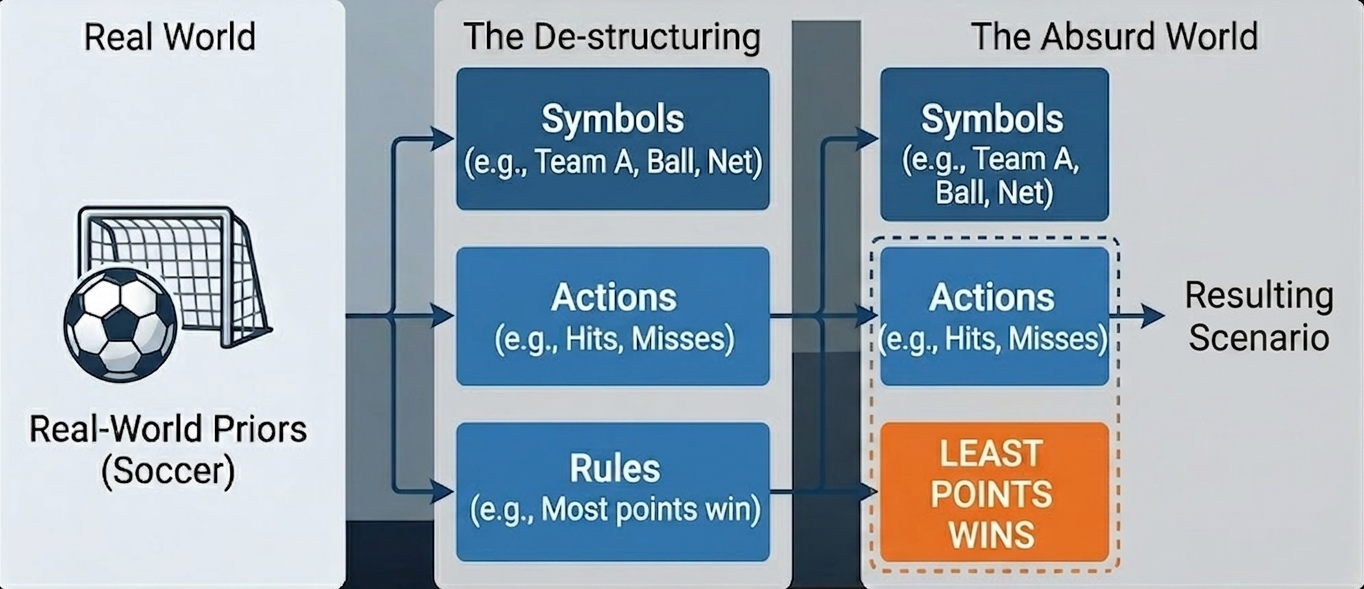}
    \caption{Decomposition of a real-world scenario into symbolic primitives.}
    \label{fig:decompose}
\end{figure}

Figure~\ref{fig:decompose} shows how we decompose the symbols, actions, and rules. In our experiments, each modified ruleset defines a controlled world model. We generate match commentaries that follow these rules and ask LLMs to determine the winner from text alone. Because the task requires only counting a small number of hits or misses, it is trivial for humans, making it an ideal testbed for evaluating how LLMs handle changes in symbols, actions, and rules without increasing task complexity.

\subsection{Rulesets \& Matches}


We first define the \textbf{REAL} ruleset, which mirrors conventional soccer scoring in a simplified form. In this setting, a match played between two teams of players. The game consists of five matches; in each match, both teams shoot the ball at the net. When one team shoots, the other team defends the net. If the team that is shooting hits the net, their score increases by 1. The team with the highest score wins. 

Every other ruleset is a systematic variation of the \textbf{REAL} ruleset. For instance, in the \textbf{MISSING} ruleset, the scoring mechanism is inverted: a team's score increases by 1 when they \textit{miss} the net. This seemingly minor modification fundamentally changes the outcome while preserving the logical steps to complete the task. Similarly, the \textbf{SWITCH} ruleset reverses the roles of the game objects: teams shoot \textit{a net into a ball} rather than a ball into a net. This variation tests whether models can adapt when symbolic roles are exchanged. 

Changes to scoring mechanisms (\textbf{MISSING}, \textbf{LEAST}, \textbf{ICE CREAM}, \textbf{CAR}) test whether models can track modified rules while maintaining logical consistency. Changes to object roles (\textbf{SWITCH}) test whether models can reason flexibly when familiar symbols are reassigned. Combined variations (\textbf{MISS \& SWITCH}) test whether models can simultaneously handle multiple changes from their prior knowledge. Table~\ref{tab:rulesets-1} summarizes all the variations we tested.

\begin{table}[H]
\caption{Ruleset variations used in Absurd Soccer experiments. Each ruleset modifies the \textbf{REAL} ruleset in specific ways.}
\label{tab:rulesets-1}
\vskip 0.15in
\begin{center}
\begin{small}
\begin{sc}
\begin{tabularx}{\linewidth}{@{}>{\raggedright\arraybackslash}p{0.23\linewidth}>{\raggedright\arraybackslash}X>{\raggedright\arraybackslash}p{0.22\linewidth}@{}}
\toprule
\textbf{Ruleset} & \textbf{Modification} & \textbf{Change Type} \\
\midrule
MISSING & Score increases by 1 when a shot misses the net & Actions \\
LEAST & Team with the lowest score wins & Rules \\
ICE CREAM & Teams earn ice cream instead of points when hitting the net & Symbols \\
CAR & Teams earn cars instead of points when hitting the net &  Symbols \\
SWITCH & Teams shoot a net into a ball (roles reversed) & Symbols \\
MISS \& SWITCH & Combination of MISSING and SWITCH & Actions \& Symbols \\
\bottomrule
\end{tabularx}
\end{sc}
\end{small}
\end{center}
\vskip -0.1in
\end{table}

\section{Related Work}
\label{sec:related}
Recent work shows that large language models exhibit significant reasoning and generalization failures even on tasks that are simple for humans. These failures have been attributed to memory limits, architectural constraints, instruction-following biases, and structural sensitivity. Our work isolates rule execution failures in small, fully specified settings where memory and capacity are not limiting. We categorized the recent works based on the limitations they are focused on and the testing mechanisms.

\subsection{Reasoning Limits Beyond Memory and High Complexity}

Existing studies link LLM reasoning failures to memory and architectural limits, with ~\cite{abbe2024far} showing breakdowns beyond effective context horizons. Our work instead examines failures in small tasks where memory is not limiting. Beyond memory constraints, ~\cite{yadav2025hopskipoverthinkdiagnosing} show that language models reason correctly on simple tasks but break down as task complexity increases. Similarly, ~\cite{shojaee2025illusionthinkingunderstandingstrengths} shows even when memory is sufficient, performance still degrades as tasks become more complex. In contrast, our task is intentionally simple and not memory-bound, allowing us to isolate reasoning failures independent of task complexity.

\subsection{Instruction Following, Priors, and Rule Execution}

Studies show that LLM behavior is strongly shaped by pretraining-induced priors, while existing evaluations struggle to isolate failures of rule execution. Instruction-following studies show that RLHF improves compliance yet breaks down when instructions contradict dominant train-ing patterns \cite{ouyang2022traininglanguagemodelsfollow, zhou2023limaalignment}, while related benchmarks show that models can state rules correctly but violate them during execution  \cite{ai6010012} or deviate from task requirements due to safety alignment pressures  \cite{yi2025goodbadfailurellms}. These failures persist in small rule-based tasks, where larger models rely on memorized heuristics when rules conflict with priors \cite{mckenzie2024inversescalingbiggerisnt}, with similar effects observed in TruthfulQA  \cite{lin2022truthfulqameasuringmodelsmimic} and the Reversal Curse \cite{berglund2024reversalcursellmstrained}. Prior work on in-context learning further shows that models follow structural patterns in examples rather than executing intended rules \cite{min2022rethinkingroledemonstrationsmakes, li2025llmseasilylearnreason}, and that even explicit chain-of-thought reasoning degrades under out-of-distribution queries ~\cite{zhao2026chainofthoughtreasoningllmsmirage}

Our benchmark isolates this by directly testing rule execution in small, fully specified tasks, revealing failures that arise specifically when explicit rules contradict learned domain expectations.

\subsection{Small Tasks and Rule Generalization}
Rule generalization has been evaluated in small, controlled tasks where memory is not a limiting factor. \cite{barron2025bigthinkcapacitymemorization} showed that in synthetic arithmetic tasks, smaller models extrapolate to unseen cases, while larger models memorize training examples, and that adding factual recall eliminates generalization at all model sizes. Related world-model benchmarks evaluate prediction under fixed dynamics: ~\cite{wang2024languagemodelsservetextbased} and \cite{xie2024makinglargelanguagemodels} assess LLMs as world simulators by testing their ability to predict state transitions and outcomes given fixed rules, while ~\cite{sun2024instructionfollowingevaluatinginferential} evaluate inferential rule application in factual scenarios with RuleBench. These settings keep the underlying rules unchanged. Whereas, our work instead modifies the rules themselves while maintaining task complexity constant, isolating whether models can execute explicitly stated rules when the causal structure departs from prior expectations.

\subsection{Fragility Under Minimal Structural Variation}

LLM reasoning has been shown to be highly fragile even in small, low-complexity tasks under minimal changes: adding a single irrelevant clause can cause large performance drops ~\cite{mirzadeh2025gsmsymbolicunderstandinglimitationsmathematical},  while \cite{gema2025inversescalingtesttimecompute} demonstrate that increasing test-time reasoning length can further degrade accuracy. \cite{lu2025reasoning} showed that reasoning-specialized models fail to maintain consistent solution strategies across nearly identical problems, and ~\cite{shojaee2025illusionthinkingunderstandingstrengths} found that both standard and reasoning-focused models collapse beyond modest compositional variation, even with extended reasoning traces.

~\cite{nezhurina2025alicewonderlandsimpletasks} introduced the Alice in Wonderland benchmark, showing that state-of-the-art models can exhibit complete reasoning breakdown on deliberately simple tasks under structure-preserving variations that alter only the specific values in the problem while leaving the reasoning rule unchanged. While this work establishes extreme brittleness under small perturbations, it does not isolate which components of a task drive failure. In contrast, our work explicitly decomposes tasks into symbols, actions, and rules and varies them independently, enabling controlled attribution of failures to rule changes rather than surface form or general reasoning instability.

\section{Experimental Settings}
\label{sec:experimental-settings}

In order to examine how LLMs are affected by changes in symbols, actions, and rules, we tested various models to answer questions about the rulesets of absurd soccer we described in Section \ref{sec:the-absurd-world}. To do this, we created a task, ``Determine Outcome", and asked models to complete this task under specific rulesets and prompting techniques.

\subsection{The Task}


Our work defines a single task, "Determine Outcome" (DO), to test models' reasoning capabilities for each ruleset. This task is supposed to test models to see if they can correctly determine the outcome of a game, based on a a full game commentary and a specified ruleset. 

We tested this task on a variety of models, as described in Section \ref{subsec:models}, and used two prompt techniques for this task: (a) Zero-Shot (DO-0) and (b) Few-Shot (DO-FS). 


\textbf{DO-0.}\quad Models are given a prompt generated by the ``Game Generator" as described in Section \ref{subsec:the-generator}. For every ruleset that we tested, we used the DO-0 prompting technique 100 times on each model. The performance of a model is calculated as the proportion of the 100 prompts it had correct responses to. 

\textbf{DO-FS.}\quad Models are given a prompt that contains 3 correct question-answer pairs (with ``Q:" and ``A:" preceding every question and answer, respectively), an additional question (preceded by ``Q:"), and then an ``A:", indicating that the model should answer the last question. All 4 of these questions used the same ruleset and were randomly taken from questions that were generated while testing models on the zero-shot DO-0 prompting technique, which were selected using the following process:

\begin{enumerate}
  \item For each ruleset, obtain all prompts used in the DO-0 task that used that ruleset.
  \item Calculate the number of models which correctly answered each prompt. Use this number to ``score" each prompt. 
  \item For each ruleset, using the scores, select the lowest-performing 50\% prompts (those prompts performing less than or equal to the median).
\end{enumerate}

For every ruleset that we tested, we used the DO-FS prompting technique 100 times on every model. The performance of a model is calculated as the proportion of the 100 prompts it had correct responses to. 






\subsection{Variation Generator}\label{subsec:the-generator}


To generate Absurd World prompts, we used the following prompt template and game generation symbols. 

\begin{promptbox}{Game Generation Symbols}\label{generation-symbols}
\vspace{0.3cm}
    \textbf{Symbol\_Player:} [team, player] \\\textbf{Symbol\_Player\_1:} [Team A, Team 1, Adam] \\
    \textbf{Symbol\_Player\_2:} [Team B, Team 2, Bob] \\
    \textbf{Symbol\_Ball:} [player, ball, net] \\
    \textbf{Symbol\_Net:} [net, ball] \\
    \textbf{Symbol\_Action:} [hits, misses] \\
    \textbf{Symbol\_Comparator:} [most, least] \\
    \textbf{Symbol\_Score:} [score, point, car, ice cream]  \\
\end{promptbox}

\begin{promptbox}{Game Generation Prompt Template}\label{generation-template}
 Absurd soccer is played by two \textbf{[Symbol\_Player]}s. In one match of this game, each \textbf{[Symbol\_Player]} takes a turn to shoot a \textbf{[Symbol\_Ball]} five times at a \textbf{[Symbol\_Net]}. A \textbf{[Symbol\_Player]} can shoot only once in a match. When one \textbf{[Symbol\_Player]} shoots, the other team defends the \textbf{[Symbol\_net]}. If the \textbf{[Symbol\_Player]} that makes the shot \textbf{[Symbol\_Action]} the \textbf{[Symbol\_Net]}, that \textbf{[Symbol\_Player]} gets one \textbf{[Symbol\_Score]}. At the end of the match, the team having the \textbf{[Symbol\_Comparator]} \textbf{[Symbol\_Score]}s wins. \\

Here is the match commentary for a game of absurd soccer:

\textbf{[Symbol\_Player\_1]} shoots the \textbf{[Symbol\_Ball]} and \textbf{[HITS/MISSES]} the \textbf{[Symbol\_Net]}. \\
\textbf{[Symbol\_Player\_2]} shoots the \textbf{[Symbol\_Ball]} and \textbf{[HITS/MISSES]} the \textbf{[Symbol\_Net]}. \\
\ldots \\
\textbf{[Symbol\_Player\_1]} shoots the \textbf{[Symbol\_Ball]} and \textbf{[HITS/MISSES]} the \textbf{[Symbol\_Net]}. \\
\textbf{[Symbol\_Player\_2]} shoots the \textbf{[Symbol\_Ball]} and \textbf{[HITS/MISSES]} the \textbf{[Symbol\_Net]}.\\

Who won the game? Answer `\textbf{[Symbol\_Player\_1]}' if \textbf{[Symbol\_Player\_1]} wins, `\textbf{[Symbol\_Player\_2]}' if \textbf{[Symbol\_Player\_2]} wins, and `both \textbf{[Symbol\_Player]}s' if both \textbf{[Symbol\_Player]}s win. Please work out your reasoning process for the answer, and place your answer within two curly brackets (ex. \{\textbf{[Symbol\_player-1]}\}).
\end{promptbox}
In this generator, we randomize the symbols, actions, rules, and events only, keeping the sequence of the events intact.

\begin{figure} [H]
    \centering
    \includegraphics[width=0.75\linewidth]{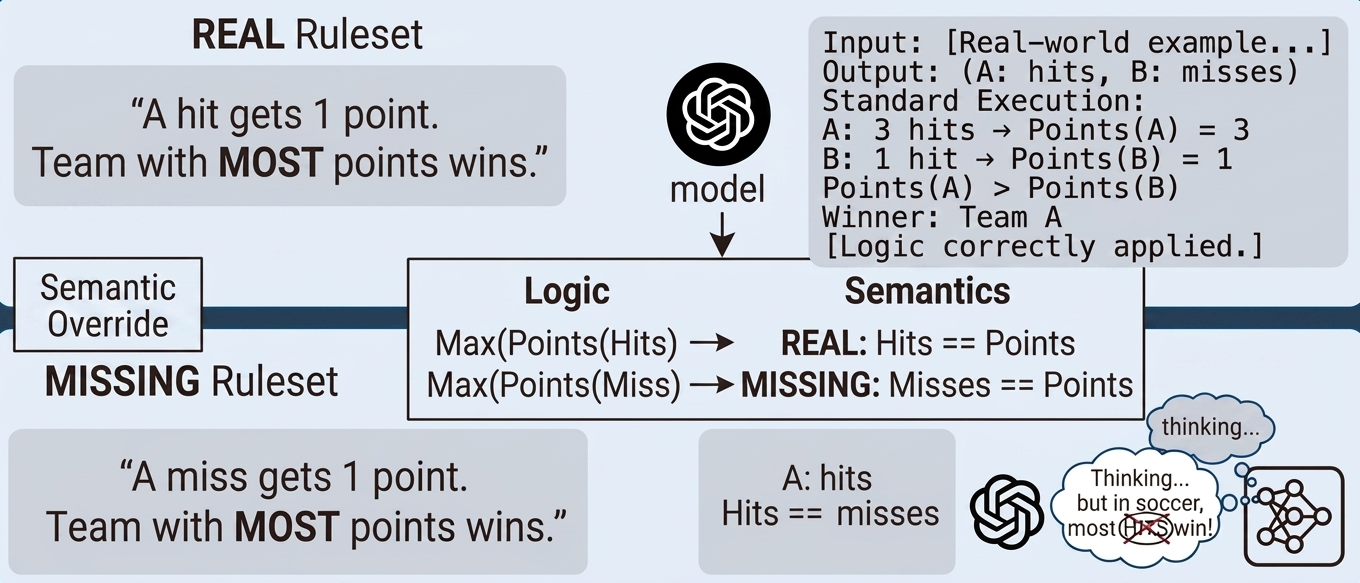}
    \caption{Semantic prior conflict in Absurd World.}
    \label{fig:example}
\end{figure}

Prompts are generated by taking the template and filling it in with the correct game generation symbols corresponding to the relevant ruleset. For instance, if we wanted to generate a prompt according to the REAL ruleset, then we would replace \textbf{[Symbol\_Ball]} with ``Ball" and \textbf{[Symbol\_Net]} with ``Net." The match commentary is then generated by randomly replacing \textbf{[HITS/MISSES]} with ``hits" or ``misses," each being equally likely. In other words, each player has the same probability of hitting as they do missing. Figure~\ref{fig:example} illustrates this semantic conflict using the REAL and MISSING rulesets. While both tasks require identical bookkeeping and winner selection, the semantic mapping between events and rewards is inverted in the absurd variant, forcing the model to override familiar domain associations in order to execute the explicitly stated rules.

\subsection{Language Models}
\label{subsec:models}

For this experiment, we tested a diverse set of LLMs from OpenRouter taken from three categories: (a) cheap non-reasoning LLMs (\$0.1-\$0.2 per million input tokens), (b) expensive non-reasoning LLMs (\$0.5-\$1.0 per million input tokens), and (c) reasoning LLMs. This categorization allows us to assess reasoning performance across a range of model sizes and architectural capabilities even when the models sizes are unknown. Models from each category were chosen based on their popularity on OpenRouter. The model names from OpenRouter are shown in Table \ref{tab:model_groups_vertical}.

\begin{table*}[t]
\caption{Models categorized by cost and reasoning specialization.}
\label{tab:model_groups_vertical}
\vskip 0.15in
\begin{center}
\begin{small}
\begin{sc}
\begin{tabular}{lcccr}
\toprule
\textbf{Category} & \textbf{Model} \\
\midrule
{Cheap} 
    & google/gemini-2.0-flash-001 \\
    & openai/gpt-4o-mini \\
    & meta-llama/llama-4-maverick-17b-128e-instruct \\
    & qwen/qwen-2.5-72b-instruct \\
    & google/gemini-2.5-flash \\
\midrule
{Expensive} 
    & anthropic/claude-3-5-haiku \\
    & nousresearch/hermes-3-llama-3.1-405b \\
    & sao10k/l3.1-euryale-70b \\
    & meta-llama/llama-3.1-405b-instruct \\
    & thedrummer/skyfall-36b-v2 \\
    & openai/gpt-4.1-mini \\
    & google/gemma-2-27b-it \\
\midrule
{Reasoning} 
    & deepseek/deepseek-r1-0528 \\
    & nvidia/llama-3.1-nemotron-ultra-253b-v1 \\
\bottomrule
\end{tabular}
\end{sc}
\end{small}
\end{center}
\vskip -0.1in
\end{table*}

\section{Results}
\label{sec:results}

\begin{figure}[hptb]
\centering
\includegraphics[width=0.4\textwidth]{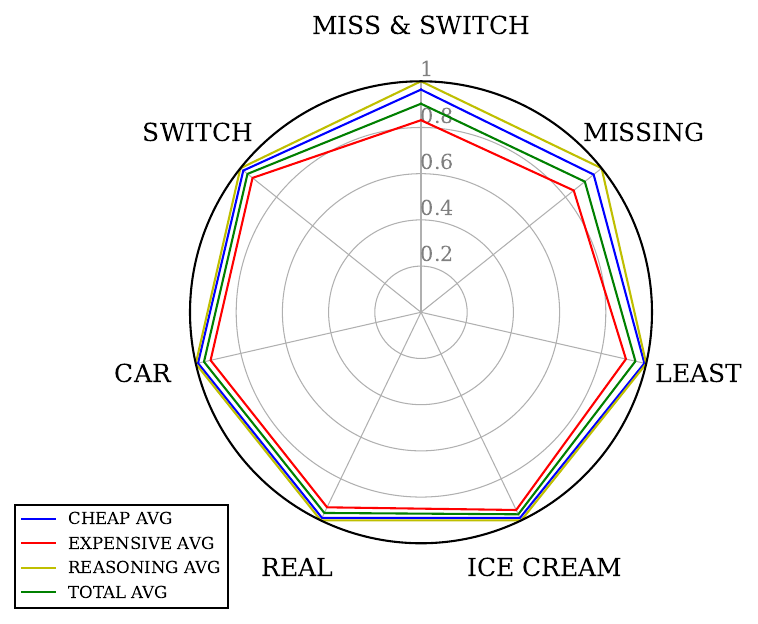}
\caption{Spider graph displaying average DO-0 scores on all tested rulesets (MISS \& SWITCH, MISSING, LEAST, ICE CREAM, REAL, CAR, SWITCH) for all model categories (CHEAP AVG, EXPENSIVE AVG, REASONING AVG), and average scores across all models (TOTAL).} 
\label{fig:cat_spider}
\end{figure}

\begin{figure*}[t]
    \centering
    \begin{minipage}{0.23\textwidth}
        \centering
        \includegraphics[width=\textwidth]{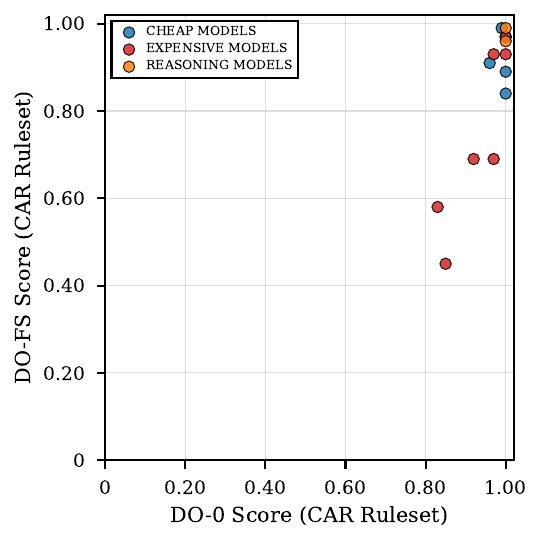}
        \small{(a) CAR}
    \end{minipage}
    \hfill
    \begin{minipage}{0.23\textwidth}
        \centering
        \includegraphics[width=\textwidth]{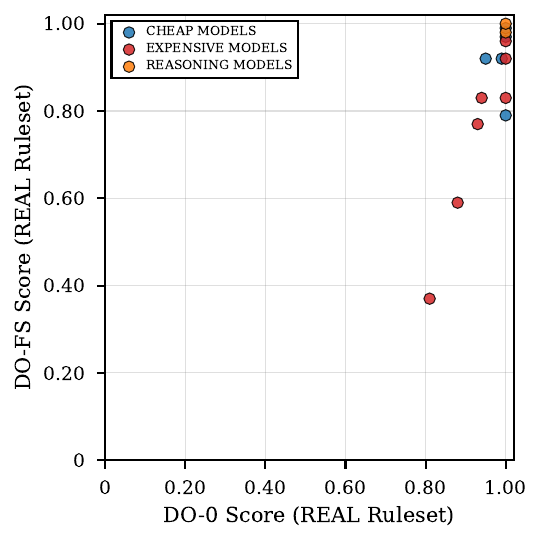}
        \small{(b) REAL}
    \end{minipage}
    \hfill
    \begin{minipage}{0.23\textwidth}
        \centering
        \includegraphics[width=\textwidth]{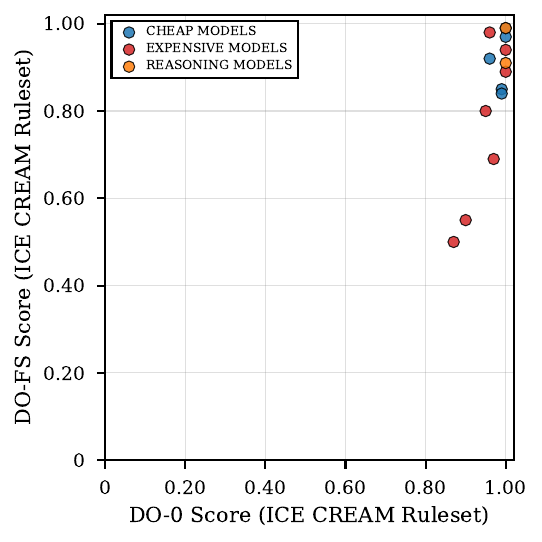}
        \small{(c) ICE CREAM}
    \end{minipage}
    \hfill
    \begin{minipage}{0.23\textwidth}
        \centering
        \includegraphics[width=\textwidth]{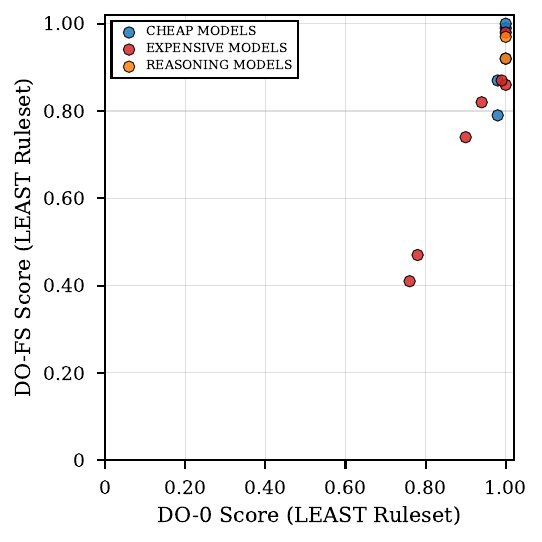}
        \small{(d) LEAST}
    \end{minipage}

    \vspace{0.5cm}

    \begin{minipage}{0.23\textwidth}
        \centering
        \includegraphics[width=\textwidth]{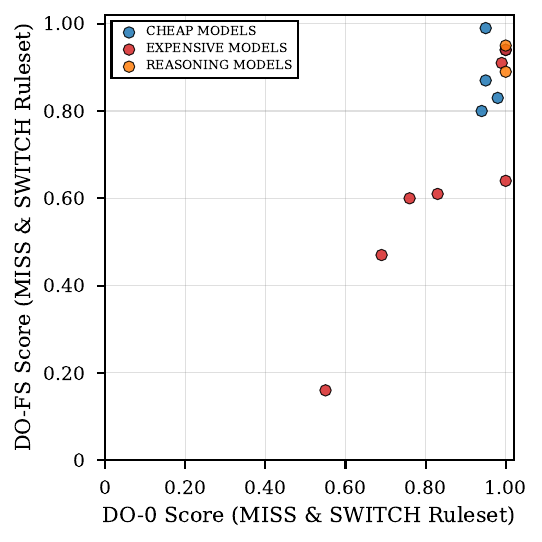}
        \small{(e) MISS \& SWITCH}
    \end{minipage}
    \quad
    \begin{minipage}{0.23\textwidth}
        \centering
        \includegraphics[width=\textwidth]{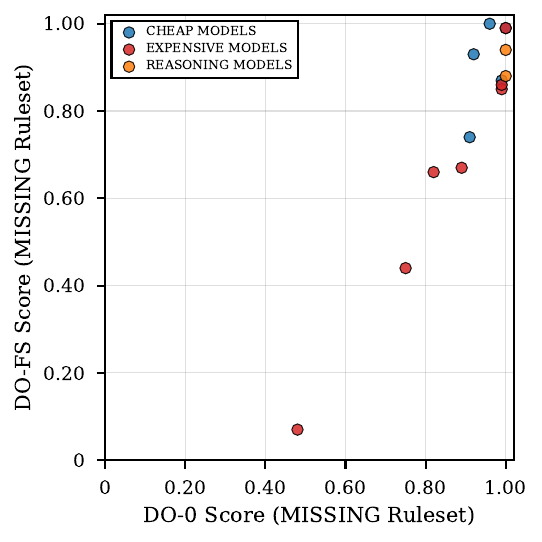}
        \small{(f) MISSING}
    \end{minipage}
    \quad
    \begin{minipage}{0.23\textwidth}
        \centering
        \includegraphics[width=\textwidth]{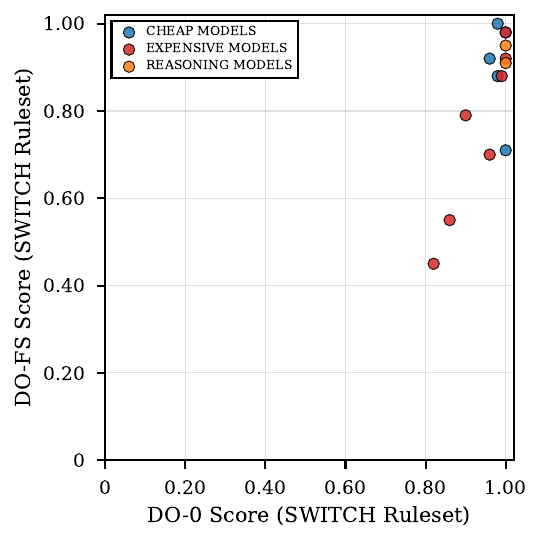}
        \small{(g) SWITCH}
    \end{minipage}

    \vspace{0.2cm}
    \caption{DO-FS scores (y-axis) compared with DO-0 scores (x-axis) for each model across seven rulesets (CAR, REAL, ICE CREAM, LEAST, MISS \& SWITCH, MISSING, SWITCH). Each dot represents a model. Models are color-coded by category: \textbf{CHEAP} (blue), \textbf{EXPENSIVE} (red), and \textbf{REASONING} (yellow).}
    \label{fig:combined_rulesets}
\end{figure*}

\begin{figure}[H]
\centering
\includegraphics[width=0.5\textwidth]{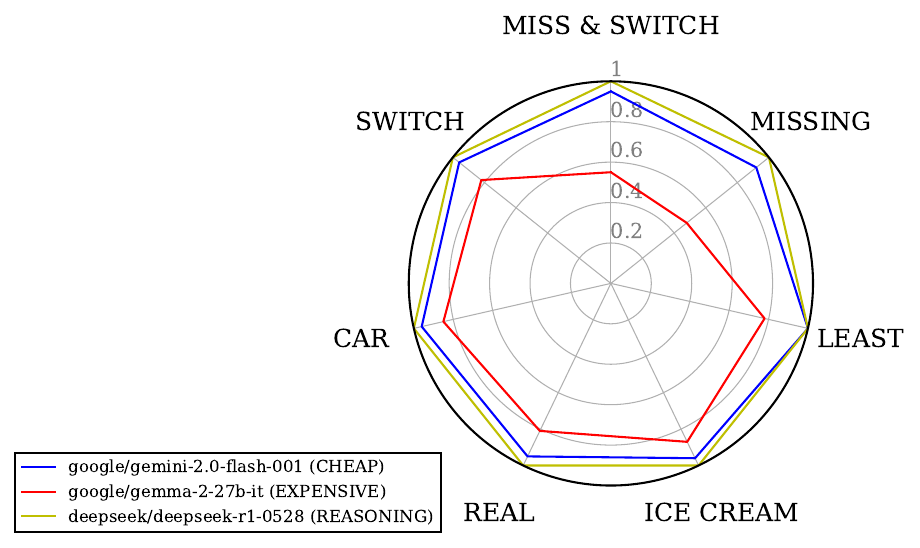}
\caption{Spider graph displaying DO-0 scores on all tested rulesets (MISS \& SWITCH, MISSING, LEAST, ICE CREAM, REAL, CAR, SWITCH) for the worst performing model in each model category (google/gemini-2.0-flash-001 (CHEAP), google/gemma-2-27b-it (EXPENSIVE), deepseek/deepseek-r1-0528 (REASONING)).}
\label{fig:worst_models_spider}
\end{figure}

After running our experiment, we calculated the score for each model on every ruleset for both prompt techniques, which was determined by the percentage of questions the model got right for that particular ruleset and prompt technique. The scores for each model category were calculated by averaging the scores of every model in that category. Three key findings are:

\begin{enumerate}
    \item Zero-shot results (DO-0) reveal that non-reasoning models (cheap and expensive), which perform around 90\% accuracy, are more sensitive to rule changes (e.g., MISSING) and more prone to making mistakes when coupled with symbol swaps (e.g., MISS \& SWITCH) than reasoning models, which have 100\% accuracy (\textit{See Fig. \ref{fig:cat_spider} and Table \ref{tab:do_results}}). We believe that this is a result of the better reasoning capabilities in reasoning LLMs. A table displaying per-model scores is located in Appendix \ref{append:per-model}. 
    
    \item Few-shot prompting consistently degrades performance rather than improving it, contrary to common expectations about in-context learning (\textit{See Table \ref{tab:dofs_results}, and Fig. \ref{fig:combined_rulesets}}). We are not sure why this is the case, but we hypothesize that this degradation indicates that these models were not familiar with the few-shot prompting technique we used. More research is needed to investigate this. Appendix \ref{append:t-test} includes a t-test to validate the statistical significance of the performance degradation. 
    \item Expensive non-reasoning models perform worse than cheap and reasoning models across both prompt techniques. As shown in Figs. \ref{fig:cat_spider} and \ref{fig:worst_models_spider}, expensive models have the lowest average accuracy across rulesets, and even the weakest expensive model underperforms the weakest models in the other categories. The entropy analysis in Appendix \ref{append:entropy} shows incorrect answers have higher entropy, but it does not give a conclusive reasoning.
\end{enumerate}

\begin{table*}[htbp]
\caption{Performance of different rulesets for DO-0. Reported numbers are average scores (higher is better).}
\label{tab:do_results}
\begin{center}
\begin{small}
\begin{sc}
\begin{tabular*}{\textwidth}{@{}lccccccc@{}}
\toprule
\rowcolor{tableHeaderBg}
\textbf{Model} & \textbf{REAL} & \textbf{MISSING} & \textbf{LEAST} & \textbf{ICE CREAM} & \textbf{CAR} & \textbf{SWITCH} & \textbf{MISS \&} \\
\rowcolor{tableHeaderBg}
\textbf{Category} & & & & & & & \textbf{SWITCH} \\
\midrule
\rowcolor{cheapRowBg}
\textbf{Cheap} & $0.988$ & $0.956$ & $0.992$ & $0.988$ & $0.990$ & $0.984$ & $0.964$ \\
\rowcolor{expensiveRowBg}
\textbf{Expensive} & $0.937$ & $0.846$ & $0.910$ & $0.950$ & $0.934$ & $0.933$ & $0.831$ \\
\rowcolor{reasoningRowBg}
\textbf{Reasoning} & $1.000$ & $1.000$ & $1.000$ & $1.000$ & $1.000$ & $1.000$ & $1.000$ \\
\bottomrule
\end{tabular*}
\end{sc}
\end{small}
\end{center}
\vskip -0.1in
\end{table*}

\begin{table*}[htbp]
\caption{Performance of different rulesets for DO-FS. Reported numbers are average scores (higher is better).}
\label{tab:dofs_results}
\begin{center}
\begin{small}
\begin{sc}
\begin{tabular*}{\textwidth}{@{}lccccccc@{}}
\toprule
\rowcolor{tableHeaderBg}
\textbf{Model} & \textbf{REAL} & \textbf{MISSING} & \textbf{LEAST} & \textbf{ICE CREAM} & \textbf{CAR} & \textbf{SWITCH} & \textbf{MISS \&} \\
\rowcolor{tableHeaderBg}
\textbf{Category} & & & & & & & \textbf{SWITCH} \\
\midrule
\rowcolor{cheapRowBg}
\textbf{Cheap} & 0.918 & 0.906 & 0.914 & 0.914 & 0.92 & 0.898 & 0.886 \\
\rowcolor{expensiveRowBg}
\textbf{Expensive} & 0.753 & 0.649 & 0.736 & 0.764 & 0.749 & 0.753 & 0.619 \\
\rowcolor{reasoningRowBg}
\textbf{Reasoning} & 0.990 & 0.910 & 0.945 & 0.950 & 0.975 & 0.930 & 0.920 \\
\bottomrule
\end{tabular*}
\end{sc}
\end{small}
\end{center}
\vskip -0.1in
\end{table*}

\section{Future Work}
\label{sec:future-work}

There are many opportunities to expand on this work. This includes testing LLMs using the Absurd World approach in other simple domains to see if there are similar patterns in the results. As a preliminary test, we have ran a small experiment to used this Absurd World approach on a different game involving opening doors. In this game, 3 players take turns opening 5 doors. Two of these doors have goats behind them. The LLM is tested on the following absurd worlds and is supposed to evaluate which player wins the game:
\begin{itemize}
    \item DO-RANDOM: the order in which the players open the door is completely randomized. The player who opens the door with a goat behind it first wins. If each player never opens a door with a goat behind it, no one wins.
    \item DO-RANDOM-LAST: same as RANDOM, but the player who opens the door with a goat behind it last wins.
\end{itemize}

We also tested models on the following prompting techniques, based on the RANDOM world:
\begin{itemize}
    \item DO-ORDER-WIN: all prompts that were tested in DO-RANDOM and which have a player winning as an outcome are selected. A random prompt from this selection is given to the models.
    \item DO-ORDER-NO-WIN: all prompts that were tested in DO-RANDOM and which have no players winning as an outcome are selected. A random prompt from this selection is given to the models.
\end{itemize}

\begin{table*}[htbp]
\caption{Performance of different rulesets for DO-RANDOM, DO-RANDOM-LAST, DO-ORDER-WIN, and DO-ORDER-NO-WIN. Reported numbers are average scores (higher is better).}
\label{tab:do_door_results}
\begin{center}
\begin{small}
\begin{sc}
\begin{tabular*}{\textwidth}{@{}lcccc@{}}
\toprule
\textbf{Model} & \textbf{DO-RANDOM} & \textbf{DO-RANDOM-LAST} & \textbf{DO-ORDER-WIN} & \textbf{DO-ORDER-} \\
\textbf{Category} & & &  & \textbf{NO-WIN} \\
\midrule
All Models & 0.989 & 0.933 & 0.996 & 0.939 \\
\bottomrule
\end{tabular*}
\end{sc}
\end{small}
\end{center}
\vskip -0.1in
\end{table*}

As shown in Table \ref{tab:do_door_results}, the average model performance on DO-RANDOM is greater than DO-RANDOM-LAST, and performance on DO-ORDER-WIN is greater than DO-ORDER-NO-WIN. This implies that models are more familiar with determining the outcome of this game if (1) the rules denote the player who finds a goat first as the winner and (2) there is a definite winner at the end of the game. 

Here are some other directions we intend to further this research:

\begin{itemize}
    \item Evaluate Absurd World approach in various domains and find the relationship between the performance in real-world tasks and their absurd counterparts. Essentially, whether an LLM performing well in the Absurd World can perform well in the real world is a necessary next step.
    \item Evaluate more advanced prompting techniques and agentic workflows. While counterintuitive, these advanced prompting techniques may have the similar phenomena as the few-shot experiments, where LLMs have worse performance in absurd worlds.
    \item More complicated tasks such as writing commentary for a game of absurd soccer based on a ruleset and outcome.
    \item Explanation of why some categories of models are worse than another in the absurd settings.
\end{itemize}

\begin{table*}[htbp]
\caption{Performance of different rulesets for DO-FS. Reported numbers are average scores (higher is better).}
\label{tab:dofs_results-1}
\begin{center}
\begin{small}
\begin{sc}
\begin{tabular*}{\textwidth}{@{}lccccccc@{}}
\toprule
\rowcolor{tableHeaderBg}
\textbf{Model} & \textbf{REAL} & \textbf{MISSING} & \textbf{LEAST} & \textbf{ICE CREAM} & \textbf{CAR} & \textbf{SWITCH} & \textbf{MISS \&} \\
\rowcolor{tableHeaderBg}
\textbf{Category} & & & & & & & \textbf{SWITCH} \\
\midrule
\rowcolor{cheapRowBg}
\textbf{Cheap} & 0.918 & 0.906 & 0.914 & 0.914 & 0.92 & 0.898 & 0.886 \\
\rowcolor{expensiveRowBg}
\textbf{Expensive} & 0.753 & 0.649 & 0.736 & 0.764 & 0.749 & 0.753 & 0.619 \\
\rowcolor{reasoningRowBg}
\textbf{Reasoning} & 0.990 & 0.910 & 0.945 & 0.950 & 0.975 & 0.930 & 0.920 \\
\bottomrule
\end{tabular*}
\end{sc}
\end{small}
\end{center}
\vskip -0.1in
\end{table*}

\section{Conclusion}

In this paper, we proposed Absurd World, a benchmarking framework for evaluating LLMs in logically coherent scenarios that deliberately violate real-world expectations, yet remain easily solvable by humans. Our results show substantial performance differences across rulesets, with particularly large failures under rule changes that directly alter scoring mechanics (MISSING and MISS \& SWITCH). We believe that this is due to expensive non-reasoning models being less confident in their responses, as displayed by our entropy analysis experiment. We further find that few-shot prompting (DO-FS) unexpectedly degrades performance relative to zero-shot prompting (DO-0), suggesting that in-context examples can interfere with correct rule execution. Together, these results expose fundamental limits in LLMs’ ability to execute explicit rules when the task’s logic deviate from real-world expectations and highlight the need for targeted benchmarks like Absurd World to guide model evaluation and selection beyond standard realism-based tasks.

\bibliography{references}

\newpage
\appendix
\section{DO-0 (zero-shot) performance per model in MISSING and MISS \& SWITCH rulesets.} \label{append:DO-0-rulesets-performance-by-model}

\begin{figure}[htbp]
\label{fig:miss_and_miss_switch}
\centering
\includegraphics[width=0.99\textwidth]{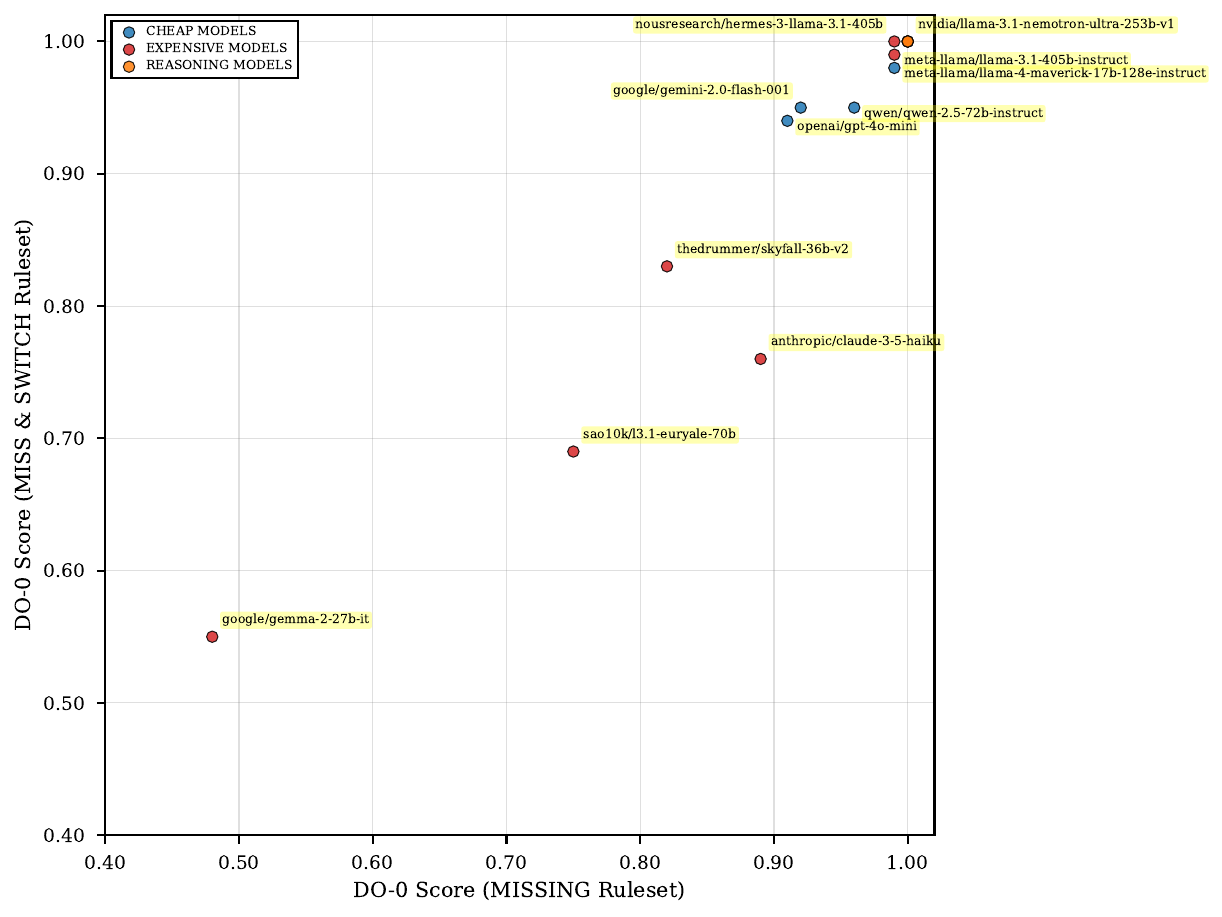}
\caption{DO-0 scores for the MISS \& SWITCH ruleset (y-axis) compared with DO-0 scores for the MISSING ruleset (x-axis). Models are labeled color-coded by category: \textbf{CHEAP} (blue), \textbf{EXPENSIVE} (red), and \textbf{REASONING} (yellow). (Models google/gemini-2.5-flash, openai/gpt-4.1-mini, and deepseek/deepseek-r1-0528 are not shown in this graph, due to have the same performance on the MISS \& SWITCH and MISSING rulesets, respectively, as nvidia/llama-3.1-nemotron-ultra-253b-v1).}
\end{figure}

\FloatBarrier

\section{Per-Model Performance Analysis}
\label{append:per-model}

\subsubsection{Zero-Shot Performance (DO-0)}

Table~\ref{tab:do0_performance} presents per-model accuracy on the DO-0 task across all seven rulesets. Reasoning models achieve perfect accuracy  across all rulesets, demonstrating robust logical reasoning capabilities. Cheap models exhibit strong performance with most achieving near-perfect baseline accuracy (mean = 0.988), though performance degrades on rule-inverted rulesets like MISSING and MISS \& SWITCH. In expensive models,  google/gemma-2-27b-it achieves only 0.48 on MISSING and sao10k/l3.1-euryale-70b drops to 0.69 on MISS \& SWITCH, despite both models maintaining reasonable performance on symbol-swap rulesets (SWITCH: 0.82 and 0.99 respectively)
\begin{table*}[t]
\centering
\caption{Per-model performance on DO-0 (Zero-Shot) task across all rulesets. Values represent accuracy scores (higher is better).}
\label{tab:do0_performance}
\resizebox{\textwidth}{!}{
\begin{tabular}{l|ccccccc}
\toprule
\rowcolor{tableHeaderBg}
\textbf{Model} & \textbf{REAL} & \textbf{MISSING} & \textbf{LEAST} & \textbf{ICE CREAM} & \textbf{CAR} & \textbf{SWITCH} & \textbf{MISS \& SWITCH} \\
\midrule
\rowcolor{cheapLabelBg}
\multicolumn{8}{l}{\textbf{Cheap Models}} \\
\midrule
\rowcolor{cheapRowBg}
google/gemini-2.0-flash-001 & \textbf{0.95} & 0.92 & 1.00 & 0.96 & 0.96 & 0.96 & 0.95 \\
\rowcolor{white}
openai/gpt-4o-mini & \textbf{0.99} & 0.91 & 0.98 & 0.99 & 1.00 & 0.98 & 0.94 \\
\rowcolor{cheapRowBg}
meta-llama/llama-4-maverick-17b-128e-instruct & \textbf{1.00} & 0.99 & 0.98 & 0.99 & 1.00 & 1.00 & 0.98 \\
\rowcolor{white}
qwen/qwen-2.5-72b-instruct & \textbf{1.00} & 0.96 & 1.00 & 1.00 & 0.99 & 0.98 & 0.95 \\
\rowcolor{cheapRowBg}
google/gemini-2.5-flash & \textbf{1.00} & 1.00 & 1.00 & 1.00 & 1.00 & 1.00 & 1.00 \\
\midrule
\rowcolor{expensiveLabelBg}
\multicolumn{8}{l}{\textbf{Expensive Models}} \\
\midrule
\rowcolor{expensiveRowBg}
anthropic/claude-3-5-haiku & \textbf{0.94} & 0.89 & 0.94 & 0.95 & 0.92 & 0.91 & 0.76 \\
\rowcolor{white}
nousresearch/hermes-3-llama-3.1-405b & \textbf{1.00} & 0.99 & 1.00 & 1.00 & 1.00 & 0.86 & 1.00 \\
\rowcolor{expensiveRowBg}
sao10k/l3.1-euryale-70b & \textbf{0.88} & 0.75 & 0.76 & 0.90 & 0.83 & 0.99 & 0.69 \\
\rowcolor{white}
meta-llama/llama-3.1-405b-instruct & \textbf{1.00} & 0.99 & 0.99 & 0.96 & 0.97 & 0.96 & 0.99 \\
\rowcolor{expensiveRowBg}
thedrummer/skyfall-36b-v2 & \textbf{0.93} & 0.82 & 0.91 & 0.97 & 0.97 & 1.00 & 0.83 \\
\rowcolor{white}
openai/gpt-4.1-mini & \textbf{1.00} & 1.00 & 0.78 & 1.00 & 1.00 & 1.00 & 1.00 \\
\rowcolor{expensiveRowBg}
google/gemma-2-27b-it & \textbf{0.81} & 0.48 & 0.78 & 0.87 & 0.85 & 0.82 & 0.55 \\
\midrule
\rowcolor{reasoningLabelBg}
\multicolumn{8}{l}{\textbf{Reasoning Models}} \\
\midrule
\rowcolor{reasoningRowBg}
deepseek/deepseek-r1-0528 & \textbf{1.00} & 1.00 & 1.00 & 1.00 & 1.00 & 1.00 & 1.00 \\
\rowcolor{white}
nvidia/llama-3.1-nemotron-ultra-253b-v1 & \textbf{1.00} & 1.00 & 1.00 & 1.00 & 1.00 & 1.00 & 1.00 \\
\bottomrule
\end{tabular}
}
\end{table*}

\subsubsection{Few-Shot Performance (DO-FS)}

Table~\ref{tab:dofs_performance} reveals systematic performance degradation under few-shot prompting across nearly all models and rulesets. Reasoning models maintain the highest accuracy but experience notable drops, with deepseek/deepseek-r1-0528 decreasing from perfect 1.00 on DO-0 to 0.88-0.98 on DO-FS depending on the ruleset. Cheap models show mixed responses, with qwen/qwen-2.5-72b-instruct maintaining near-perfect performance while others experience substantial degradation. Expensive models suffer the most severe few-shot degradation, with google/gemma-2-27b-it collapsing to 0.07 on MISSING and anthropic/claude-3-5-haiku dropping from 0.89 to 0.67 on the same ruleset. 

\begin{table*}[h]
\centering
\caption{Per-model performance on DO-FS (Few-Shot) task across all rulesets. Values represent accuracy scores (higher is better).}
\label{tab:dofs_performance}
\resizebox{\textwidth}{!}{
\begin{tabular}{l|ccccccc}
\toprule
\rowcolor{tableHeaderBg}
\textbf{Model} & \textbf{REAL} & \textbf{MISSING} & \textbf{LEAST} & \textbf{ICE CREAM} & \textbf{CAR} & \textbf{SWITCH} & \textbf{MISS \& SWITCH} \\
\midrule
\rowcolor{cheapLabelBg}
\multicolumn{8}{l}{\textbf{Cheap Models}} \\
\midrule
\rowcolor{cheapRowBg}
google/gemini-2.0-flash-001 & \textbf{0.92} & 0.93 & 0.92 & 0.92 & 0.91 & 0.92 & 0.87 \\
\rowcolor{white}
openai/gpt-4o-mini & \textbf{0.92} & 0.74 & 0.87 & 0.85 & 0.89 & 0.88 & 0.80 \\
\rowcolor{cheapRowBg}
meta-llama/llama-4-maverick-17b-128e-instruct & \textbf{0.79} & 0.87 & 0.79 & 0.84 & 0.84 & 0.71 & 0.83 \\
\rowcolor{white}
qwen/qwen-2.5-72b-instruct & \textbf{0.97} & 1.00 & 0.99 & 0.97 & 0.99 & 1.00 & 0.99 \\
\rowcolor{cheapRowBg}
google/gemini-2.5-flash & \textbf{0.99} & 0.99 & 1.00 & 0.99 & 0.97 & 0.98 & 0.94 \\
\midrule
\rowcolor{expensiveLabelBg}
\multicolumn{8}{l}{\textbf{Expensive Models}} \\
\midrule
\rowcolor{expensiveRowBg}
anthropic/claude-3-5-haiku & \textbf{0.83} & 0.67 & 0.82 & 0.80 & 0.69 & 0.79 & 0.60 \\
\rowcolor{white}
nousresearch/hermes-3-llama-3.1-405b & \textbf{0.83} & 0.85 & 0.86 & 0.89 & 0.93 & 0.92 & 0.64 \\
\rowcolor{expensiveRowBg}
sao10k/l3.1-euryale-70b & \textbf{0.59} & 0.44 & 0.41 & 0.55 & 0.58 & 0.55 & 0.47 \\
\rowcolor{white}
meta-llama/llama-3.1-405b-instruct & \textbf{0.92} & 0.86 & 0.87 & 0.98 & 0.93 & 0.88 & 0.91 \\
\rowcolor{expensiveRowBg}
thedrummer/skyfall-36b-v2 & \textbf{0.77} & 0.66 & 0.74 & 0.69 & 0.69 & 0.70 & 0.61 \\
\rowcolor{white}
openai/gpt-4.1-mini & \textbf{0.96} & 0.99 & 0.98 & 0.94 & 0.97 & 0.98 & 0.94 \\
\rowcolor{expensiveRowBg}
google/gemma-2-27b-it & \textbf{0.37} & 0.07 & 0.47 & 0.50 & 0.45 & 0.45 & 0.16 \\
\midrule
\rowcolor{reasoningLabelBg}
\multicolumn{8}{l}{\textbf{Reasoning Models}} \\
\midrule
\rowcolor{reasoningRowBg}
deepseek/deepseek-r1-0528 & \textbf{0.98} & 0.88 & 0.92 & 0.91 & 0.96 & 0.91 & 0.89 \\
\rowcolor{white}
nvidia/llama-3.1-nemotron-ultra-253b-v1 & \textbf{1.00} & 0.94 & 0.97 & 0.99 & 0.99 & 0.95 & 0.95 \\
\bottomrule
\end{tabular}
}
\end{table*}

\FloatBarrier

\section{T-test for DO-FS and DO-0 techniques}
\label{append:t-test}

\textbf{Assumption (H$_1$)}: DO-FS (few-shot) is better than DO-0(zero-shot). 
As we found, the scores for the few-shot technique are worse than zero-shots, we tested for statistical significance using t-test. The following hypothesis tests whether zero-shot is better than few-shot with the absurd soccer games. In this hypothesis $\mu_{\text{DO-0}}$ represents the average performance of DO-0 and $\mu_{\text{DO-FS}}$ represents the average performance of DO-FS.  

\begin{equation}
    H_1: \mu_{\text{DO-0}} - \mu_{\text{DO-FS}} > 0
\end{equation}

Table~\ref{tab:hypothesis_results} shows p-values for all the rulesets (including the real soccer), which are all highly significant. Moreover, the difference between zero-shot and few-shot increases with absurd world models.

\begin{table*}
\centering
\caption{Hypothesis Testing Results Across Seven Rulesets}
\label{tab:hypothesis_results}
\begin{tabular}{lcccccc}
\toprule
\textbf{Ruleset} & \boldmath$\mu_{\text{DO-0}}$ & \boldmath$\mu_{\text{DO-FS}}$ & \textbf{Difference = $\mu_{\text{DO-0}} - \mu_{\text{DO-FS}}$} & \textbf{t-statistic} & \textbf{p-value} \\
\midrule
\textbf{REAL} & 0.964 & 0.846 & +0.1180 & 4.001 & 0.00058 \\
\textbf{MISSING} & 0.907 & 0.778 & +0.1290 & 4.123 & 0.00045 \\
\textbf{LEAST} & 0.952 & 0.829 & +0.1230 & 4.326 & 0.00030 \\
\textbf{ICE CREAM} & 0.971 & 0.844 & +0.1270 & 4.765 & 0.00013 \\
\textbf{CAR} & 0.963 & 0.842 & +0.1210 & 3.951 & 0.00064 \\
\textbf{SWITCH} & 0.961 & 0.830 & +0.1310 & 4.252 & 0.00035 \\
\textbf{MISS \& SWITCH} & 0.903 & 0.757 & +0.1460 & 4.412 & 0.00025 \\
\midrule
\textbf{AVERAGE} & \textbf{0.946} & \textbf{0.818} & \textbf{+0.1279} & \textbf{4.261} & \textbf{0.00039} \\
\bottomrule
\end{tabular}
\end{table*}
The Difference column shows that every single number is positive. The smallest difference is +0.0800 (Ice Cream). The largest is +0.1020 (Miss \& Switch). The average is +0.0903.
Seven out of seven rulesets show that DO-0 scores are higher than DO-FS.



\section{Entropy Analysis}\label{append:entropy}

Another question we'd like to answer is why some models perform worse than others by analyzing the average entropy of responses from the models listed in table \ref{tab:entropy_models}. We used models from OpenAI and Google, since they provide information on the log-probabilities for the output tokens in their LLMs. We tested these models on the DO-0 and DO-FS tasks, but also recorded an additional entropy metric for each of the models' responses.

\begin{table}[t]
\caption{OpenAI and Google models used in entropy analysis experiment}
\vskip 0.15in
\begin{center}
\begin{small}
\begin{sc}
\begin{tabular}{lcccr}
\toprule
\textbf{{Provider}} & \textbf{Model} \\
\midrule
{OpenAI} 
    & gpt-4o-mini-2024-07-18 \\
    & gpt-4o-mini \\
    & gpt-3.5-turbo \\
\midrule
{Google} 
    & gemini-2.5-flash-lite \\
    & gemini-2.0-flash-001 \\
    & gemini-3-flash-preview \\
\bottomrule
\end{tabular}
\end{sc}
\end{small}
\end{center}
\vskip -0.1in
\label{tab:entropy_models}
\end{table}

\begin{figure}[H]
    \centering
    \begin{minipage}{0.46\textwidth}
        \small{DO-0}
        \centering
        \includegraphics[width=\textwidth]{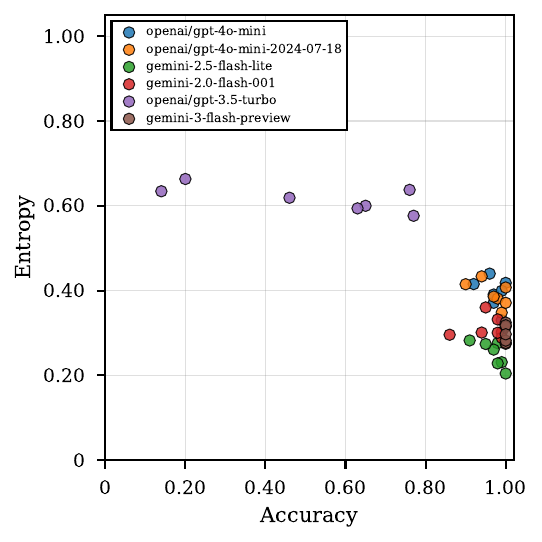}
    \end{minipage}
    \hfill
    \begin{minipage}{0.46\textwidth}
        \small{DO-FS}
        \centering
        \includegraphics[width=\textwidth]{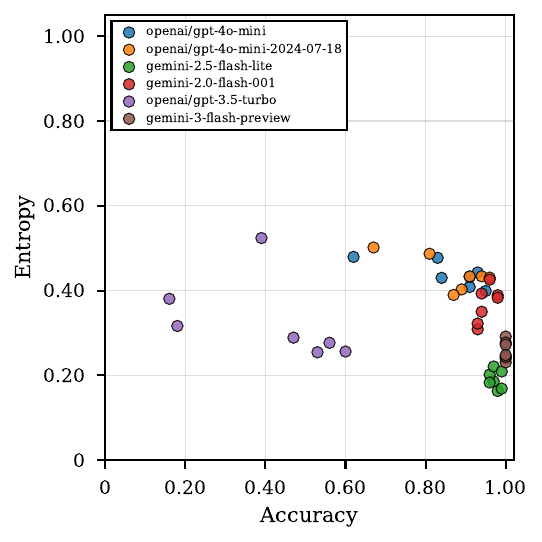}
    \end{minipage}

    \vspace{0.5cm}

    \begin{minipage}{0.46\textwidth}
        \centering
        \includegraphics[width=\textwidth]{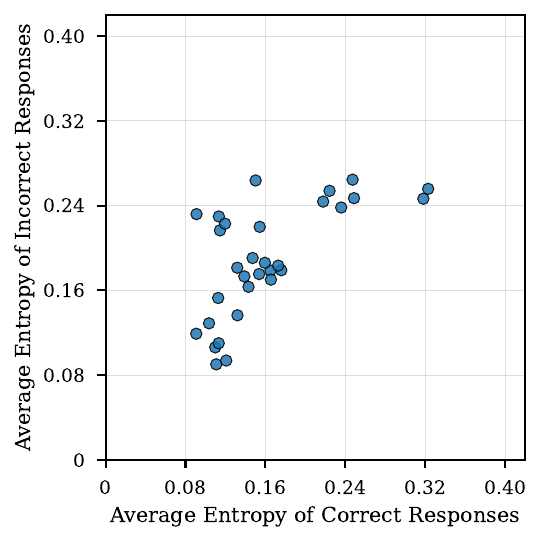}
    \end{minipage}
    \hfill
    \begin{minipage}{0.46\textwidth}
        \centering
        \includegraphics[width=\textwidth]{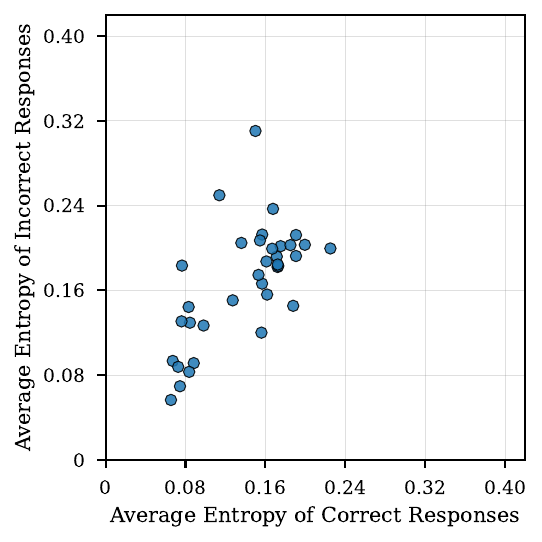}
    \end{minipage}

    \caption{\textbf{TOP HALF:} DO-0 (left) and DO-FS (right) scores compared with average entropy. Each dot represents a model-ruleset pair (that is, the results for a particular model on a particular ruleset), color-coded by model. \textbf{BOTTOM HALF:} Average entropy in incorrect answers compared with average entropy in correct answers for DO-0 (left) and DO-FS (right). Each dot represents a model-ruleset pair. (Model-ruleset pairs in which the model got all of the questions right are not shown).}
    \label{fig:combined_entropy}
\end{figure}
As shown in the top-half of Figure \ref{fig:combined_entropy}, performance on DO-0 and DO-FS is negatively correlated with entropy, although the correlation is weaker for DO-FS. In the bottom half of Figure \ref{fig:combined_entropy}, the entropy of incorrect responses is often greater than that of correct responses. These results imply that models are less confident in their answers when they answer a question incorrectly, which does not establish any bias towards the real-world priors; they can be wrong by chance. Further research is necessary to understand why advanced prompting techniques are worse in absurd worlds.

\end{document}